\documentclass[conference]{IEEEtran}
\IEEEoverridecommandlockouts
\usepackage{cite}
\usepackage{amsmath,amssymb,amsfonts}
\usepackage{algorithmic}
\usepackage{graphicx}
\usepackage{textcomp}
\usepackage{xcolor}

\usepackage{times}
\usepackage{soul}
\usepackage{url}
\usepackage[hidelinks]{hyperref}
\usepackage[utf8]{inputenc}
\usepackage[small]{caption}
\usepackage{graphicx}
\usepackage{amsmath}
\usepackage{booktabs}
\usepackage{algorithm}
\usepackage{algorithmic}
\usepackage{amsmath} 
\usepackage{amsfonts}
\usepackage{subcaption} 
\usepackage{multirow}
\usepackage{booktabs}
\usepackage{hyperref}       
\usepackage{color}

\def\BibTeX{{\rm B\kern-.05em{\sc i\kern-.025em b}\kern-.08em
    T\kern-.1667em\lower.7ex\hbox{E}\kern-.125emX}}
    
\DeclareMathOperator{\x}{\mathbf{x}}
\DeclareMathOperator{\X}{\mathbf{X}}
\DeclareMathOperator{\Y}{\mathbf{Y}}
\DeclareMathOperator{\z}{\mathbf{z}}

\begin{document}

\title{\Huge Performing Co-Membership Attacks Against \\Deep Generative Models}
\author{Kin Sum Liu $^{1}$\quad Chaowei Xiao $^{2}$  \quad  Bo Li  $^{3}$  \quad Jie Gao $^{1}$\\
$^1$ Stony Brook University
$^2$ University of Michigan, Ann Arbor
$^3$ UIUC\\
kiliu@cs.stonybrook.edu, xiaocw@umich.edu, lbo@illinois.edu, jgao@cs.stonybrook.edu

}

\maketitle

\begin{abstract}
In this paper we propose a new membership attack method called \emph{co-membership attacks} against deep generative models including Variational Autoencoders (VAEs) and Generative Adversarial Networks (GANs). Specifically, membership attack aims to check whether a given instance $\x$ was used in the training data or not. A \emph{co-membership attack} checks whether the given bundle of $n$ instances were in the training, with the prior knowledge that the bundle was either entirely used in the training or none at all. Successful membership attacks can compromise the privacy of training data when the generative model is published. Our main idea is to cast membership inference of target data $\x$ as the optimization of another neural network (called the \emph{attacker network}) to search for the latent encoding to reproduce $\x$. The final reconstruction error is used directly to conclude whether $\x$ was in the training data or not. We conduct extensive experiments on a variety of datasets and generative models showing that: our attacker network outperforms prior membership attacks; \emph{co-membership attacks} can be substantially more powerful than single attacks; and VAEs are more susceptible to membership attacks compared to GANs. 
\end{abstract}


\begin{figure*}[ht]
\vskip -0.1in
\begin{center}
\centerline{\includegraphics[width=0.9\textwidth]{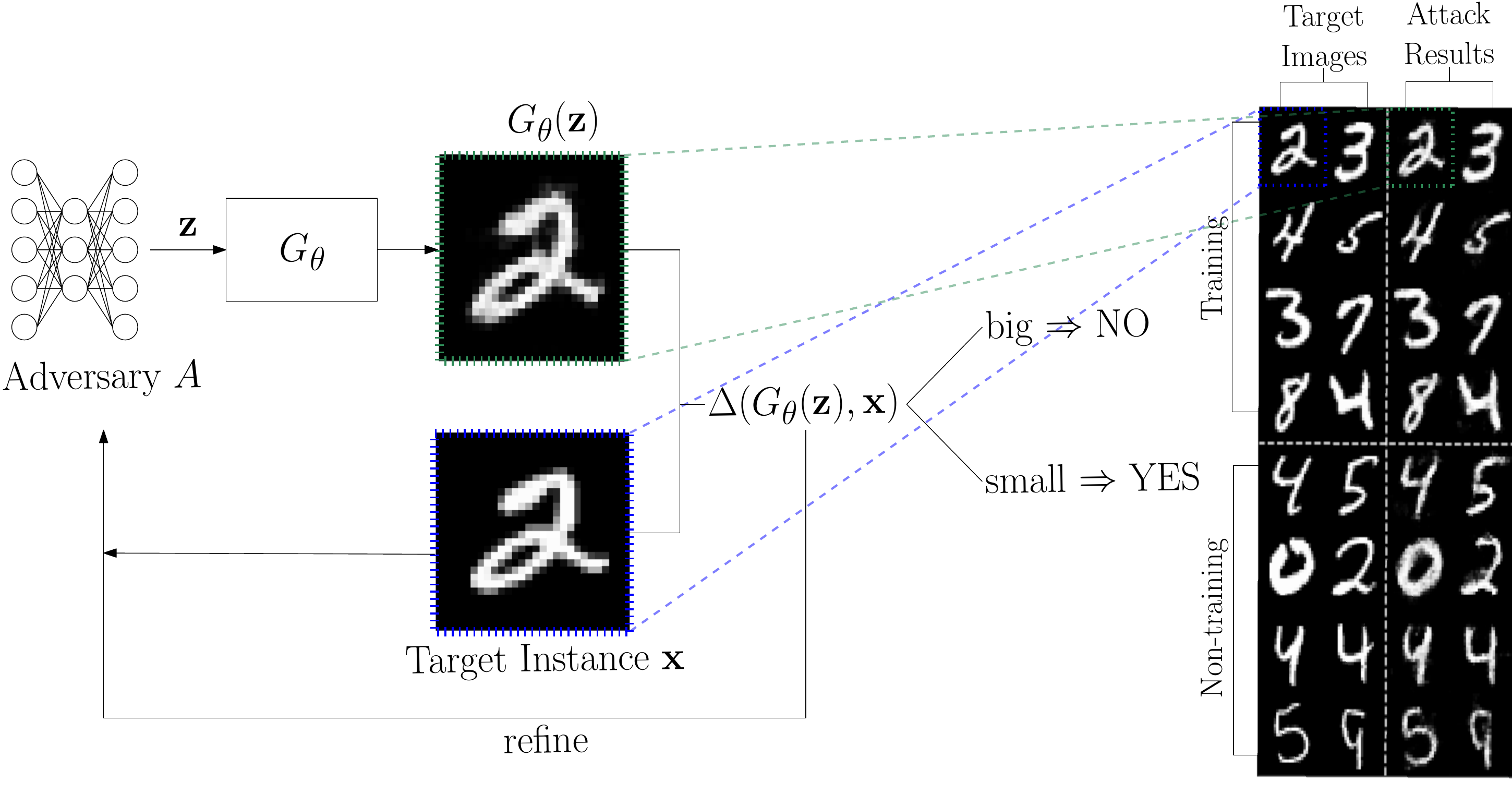}}
\caption{Membership Attack $(G_\theta, \x)$: Answer whether the target image $\x$ was used in the training data of $G_\theta$. The left side shows the attacker framework which uses the attacker loss $\Delta(G_\theta(\z),\x)$ to determine membership.}
\label{fig:membership_attack}
\vspace*{-3mm}
\end{center}
\end{figure*}

\section{Introduction}

Deep generative models have been widely studied recently as an effective method to abstract and approximate a data distribution. Two of the most commonly used ones are Variational Autoencoders (VAEs)~\cite{kingma2013auto} and Generative Adversarial Networks (GANs)~\cite{goodfellow2014generative}. 
VAEs consist of a pair of networks, the encoder and the decoder, and explicitly maximize the variational lower bound with respect to the joint generative and inference models. 
GANs take a game theoretic approach and try to achieve a good balance between two strategic agents, the generator and discriminator networks.
These generative models have been applied on complicated real-world data, with a variety of applications: in graphics, geometric modeling, and even for 
designing cryptographic primitives~\cite{abadi2016learning}. 



A successful machine learning model captures information from the training dataset. At the same time, an adversary may gain knowledge about the training dataset by interacting with the model. This information leakage is formulated by the \emph{membership inference problem} \cite{Shokri2016-lh,Yeom2017-qi}: given a model and a data point, determine whether the data point was in the training dataset of the model. This can be considered as an attack to the privacy of the training data. A successful membership attack to a machine learning model means that the privacy of the training data was not sufficiently protected when the trained model is released. 

In current literature, most membership attacks were designed for classifiers. Fredrikson et al.~\cite{usenix2014} proposed to infer sensitive features of input data by actively probing the outputs. Later, a more robust inversion attack was developed~\cite{ccs15}, in which attackers can recover part of the training set, such as human faces. Shokri et al.~\cite{shokri2016membership} tried to predict whether a data point belongs to the training set by querying the model to predict the class label of the data point. More recently, a 
GANs-based attack has been proposed to attack collaborative deep learning~\cite{hitaj2017deep} for a distributed machine learning system, in which users collaboratively train a model by sharing gradients of their locally trained models through a parameter server.
Given the fact that Google has proposed federated learning based on distributed machine learning~\cite{mcmahan2017federated} and already deployed it on mobile devices, such a GAN-based attack raises severe privacy threats. 

However, despite the wide application of generative models, so far not much has been done on analyzing the privacy vulnerabilities of such models. Unlike discriminative models such as classifiers, generative models do not provide the interactive mode of issuing queries. Thus, membership attacks against generative models require different strategies. Hayes et al.~\cite{hayes2017logan} considered attacks of GANs and offered rankings for the target data based on the probability that the instance is in the training set, but they require the knowledge of the discriminator and cannot launch attacks on individual instances. 

\smallskip\noindent\textbf{Contributions.} 
In this article, we propose a new membership attack called the \emph{co-membership attack} and new attack methods against generative models. We consider both single membership attacks and \emph{co-membership attacks}. In a single membership attack, we are given a single instance $\x$ and ask whether $\x$ was used in the training. In a co-membership attack (co-attack), we are given $n$ instances with the knowledge that either all $n$ instances were used in training or none was used. The co-membership attack model happens in many real world scenarios, as an user often contributes multiple data points to a training task -- for example, multiple pictures uploaded from a smart phone, or multi-day data from a smart meter device.

We propose a new attack method. Given a generative model $G$ with a target $\x$ (or $n$ targets in a co-attack), we 
optimize another neural network $A$ to produce an input to the generative network such that the generated output nearly matches $\x$ (or reproduces each of the $n$ targets). If we are able to reproduce elements in the training data but unable to reproduce samples not in the training data, a membership attack is successful. Notice that this is completely unsupervised and the attacker network $A$ starts with random values. 
Trying to recover the input to a GAN in order to match a given output was also studied in DefenseGAN~\cite{defenseGAN}, in which the input variable $\z$ to GANs is taken as trainable variables and gradient descent is used to minimize the difference of the generated data and target data $\x$. Compared to this technique of directly searching in latent space, our method works well with co-membership attacks, as knowledge from multiple target instances could be extracted and shared through the attacker network $A$ while the method in DefenseGAN cannot share information across different target instances. 

We evaluated our new attack methods for both VAEs and GANs on a variety of data sets (MNIST, CelebA, and ChestX-ray8) with a suite of settings (different number of training data, different architectures, training iterations, etc). Our observations are:
\begin{itemize}
    \item Co-membership attacks are more powerful than single attacks:  co-membership attacks are necessary to achieve success when generative models are trained with large datasets.
    \item Our neural network based attacker is more powerful than other optimization methods using gradient descent directly in the latent space.
    \item VAEs are more susceptible to membership attacks compared to GANs in general. 
\end{itemize}

In addition, we also discuss the success of membership attack with a few other quality measures for generative models.  
Generalization is arguably the most important objective for a machine learning model, in which the model produces useful results not only on the training data but also on data that the model has not seen. For supervised learning, the generalization error is the 
difference between classification error on the training set and error on the underlying joint probability distribution. 
For generative models, there has not been a standard measure of generalization. With a similar intuition, we measure the \emph{generalization gap} by the \emph{difference of the reconstruction error} by the membership attacker on the training data and on the test data. 
Lastly, \emph{diversity} describes whether the generator can produce different samples~\cite{arora18GAN}.
We use a measure called \emph{dispersion} to evaluate the diversity qualitatively. 
We show that empirically, dispersion, generalization error, and success rate of membership attack are all closely correlated. 

\section{Membership Attacks Against Deep Generative Models}\label{sec:membership}
In this section, we introduce an efficient and unsupervised membership attack against different deep generative models, and show that the generative models can indeed reveal the information about the training data.

\newcommand{\real}{\text{real}}

\smallskip\noindent\textbf{Attacking GANs.} 
Let $\x$ be a data instance with dimension $d$. $\mu_{\real}$ is the distribution from which $\x$ is sampled from. So $\x \sim \mu_{\real}$. The objective of a GAN \cite{goodfellow2014generative} is to learn the distribution $\mu_{\real}$; practically GANs can be used to generate new samples $\{ \x_1, \x_2, \ldots\}$ to approximate the distribution. To achieve the goal, GANs consist of two components, the generator ($G_\theta$) and the discriminator ($D_\phi$). $G_\theta$ produces a sample and $D_\phi$ tries to distinguish whether the sample is from the output distribution of generator $\mu_{G_\theta}$, where $\theta$ and $\phi$ are neural network parameters.
To generate a new data point $\x$, \emph{generator} $G_\theta$ takes a random $k$-dimensional vector $\z$ and returns $\x = G_\theta(\z)$.
The \emph{discriminator} $D_\phi$ is a function $\mathbb{R}^d \to [0,1]$. The output of $D_\phi(\x)$ is interpreted as the probability for the data $\x$ to be drawn from $\mu_{\real}$.
The objective of a GAN is 
$$ \min_{\theta} \max_{\phi} \mathbb{E}_{\x\sim\mu_{\real}} [\log D_\phi(\x)] + \mathbb{E}_{\x\sim \mu_{G_\theta}}[\log (1-D_\phi(\x))].$$
In Wasserstein GAN (WGANs) \cite{arjovsky2017wasserstein}, the measuring function is the identity function instead of the $\log$ function in which the resultant objective is 
$$ \min_{\theta} \max_{\phi} \mathbb{E}_{\x\sim\mu_{\real}} [ D_\phi(\x)] - \mathbb{E}_{\x\sim \mu_{G_\theta}}[D_\phi(\x)].$$ In the paper and our implementation, the Lipschitz constraint on the critic is implemented with weight clipping. 
Empirically, WGANs are observed to behave better than the vanilla counterpart so we use WGANs in our experiment. 


To perform a membership attack against the generator $G_\theta$ of a trained GAN for a given target instance $\x$, we propose to introduce an attacker $A$ to synthesize a seed for $G_\theta$ so as to generate an instance close to $\x$.
The pipeline of the membership attack is shown in Figure~\ref{fig:membership_attack}. 
Specifically, the attacker is a neural network $A_\gamma$, parameterized by $\gamma$, which takes $\x$ as input and maps from $\mathbb{R}^d \to \mathbb{R}^k$. Here $k$ is the input dimension of the generator. The objective of the attacker is to minimize a distance (reconstruction loss for the attacker) between the data point $\x$ and the generated data $G_\theta(A_\gamma(\x))$: 
\begin{equation}\label{eqn:attackerlossgan}
    \min_{\gamma} \Delta(\x, G_\theta(A_\gamma(\x))).
\end{equation}

The result of the optimization problem \ref{eqn:attackerlossgan} is used directly to determine the membership of $\x$. Intuitively, smaller reconstruction error indicates that $\x$ is more likely to be from the training data. Note that the parameter $\gamma$ of the attacker network $A$ is randomly initialized for a new attack so it does not require any pre-training before performing this attack.

L2-distance is taken as our distance function $\Delta(\cdot,\cdot)$ throughout the paper. We would like to remark that our proposed attack method is not specific to L2-distance. For different datasets, other application oriented metrics might be used.


    

\newcommand{\KL}{\text{KL}}

\smallskip\noindent\textbf{Attacking VAEs.} 
VAEs~\cite{Kingma2013-xj} are good at changing or exploring variations on existing data in a desired, specific direction. 
VAEs consist of a pair of connected networks, an encoder and a decoder. 
The \emph{encoder} $q_\phi$ takes input data and outputs two vectors: a vector of means $\mu$, and a vector of standard deviations $\sigma$. With re-parameterization, $\z=\mu+\sigma \epsilon$ is obtained by sampling $\epsilon \sim \mathcal{N}(0,1)$. The \emph{decoder} $g_\theta$ takes $\z$ and generates a data point $g_\theta(\z)$. The entire network is trained with the objective
$$\min_{\theta,\phi} -\mathbb{E}_{q_\phi(\z|\x)}[\log(p(\x|g_\theta(\z)))] + \KL(q_\phi(\z|\x)	\parallel p(\z)),$$
where $\KL(p || q)$ is the KL-divergence of distributions $p, q$.

Similarly when conducting membership attack against VAEs, we search for a particular $\z$ that can reproduce the target image when $\z$ is fed to the decoder $g_\theta$. The objective of the attacker is again:

\begin{equation}\label{eqn:attackerlossvae}
    \min_{\gamma} \Delta(\x, g_\theta(A_\gamma(\x))).
\end{equation}

In both optimization objectives \ref{eqn:attackerlossgan} and \ref{eqn:attackerlossvae}, the attacker network is trying to invert the generator. The generator takes a seed $\z$ and outputs $\x$, and the attacker takes $\x$ and looks for $\z$. In VAE, if the compact representation of the encoder has the same dimension as the seed, this implies that the attacker $A_{\gamma}$ is very similar to the encoder. 

\smallskip\noindent\textbf{Single Attack v.s. Co-Attack.} The attack framework introduced is able to launch membership attacks against a single target by optimizing an instance of attacker network $A_{\gamma}$. This is called a \emph{single attack}. If the attacker has more information about several target instances (for example, the target instances are known to be either all from training or non-training data), we can \emph{optimize} one single $\gamma$ on multiple attack instances at the same time instead of initializing a new $\gamma$ for each target instance. The information of multiple instances will be fused together to guide an attacker network $A_{\gamma}$. This is termed as a \emph{co-attack with strength $n$} if $n$ target instances are handled together with the prior knowledge that they have the same membership status. The new attacker loss is defined as the average of the reconstruction loss for the $n$ instances.
So the objective of such co-attacker is:
$$\min_{\gamma} \frac{1}{n} \sum_{i}^{n} \Delta(\x_i, G(A_\gamma(\x_i))).$$

Without modeling the attacker as a neural network, such a co-attack will be difficult. We observe that in the experiments, the proposed co-attack is significantly more successful across models and datasets when $n$ increases. This shows the efficiency of co-attackers to leverage such information.



\smallskip\noindent\textbf{White-box v.s. Black-box.} Since our attacker computes the minimum reconstruction error in Equation~\ref{eqn:attackerlossgan} to determine the membership label of an instance, it requires the gradient for optimization. If the internal structure (weights and architecture for back-propagation) of the generator (decoder) $G$ is exposed (i.e., in a white box attack), we can compute an analytical gradient of the distance w.r.t. $\gamma$. Otherwise, we need to use finite-difference to approximate the gradient: $ l'(\gamma)_i= \frac{l(\gamma+\mathbf{e}_i)-l(\gamma)}{\mathbf{e}_i}$ where $l(\gamma) = \Delta(\x, G(A_\gamma(\x)))$. Then the optimization of the attacker requires more black-box accesses to the generator (decoder). 
In this work, we focus on the white-box setting to explore what a powerful adversary can do based on the Kerckhoffs's principle~\cite{shannon1949communication} to better motivate defense methods. But note that the (both white-box and black-box) adversary has no information about the composition of training data, assumption of the underlying distribution and details of the training process.

\begin{table*}
  \caption{Evaluation of membership attack on the 
  MNIST (trained with 60,000 images) 
  , CelebA (trained with 500 contributors and 25 images each) 
  , ChestX-ray8 (trained with 100 contributors and 10 images each) 
  measured by AUC of ROC}
  \label{attack-table}
  \centering
  \begin{tabular}{lccccccccc}
    \toprule
    && \multicolumn{2}{c}{MNIST} & \multicolumn{2}{c}{CelebA} &
    \multicolumn{2}{c}{ChestX-ray8}
    \\
    \cmidrule(r){3-4} \cmidrule(r){5-6} \cmidrule(r){7-8} 
    Attack method & co-attack & WGANs & VAEs & WGANs & VAEs & Residual-WGANs & VAEs \\
    \midrule
    \multirow{1}{*}{Nearest Neighbor} 
    & - & 0.52 & 0.57 & 0.48 & 0.49 & 0.53 & 0.49 \\
    \midrule
    \multirow{1}{*}{Direct Projection}
    & - & 0.57 & 0.55 & 0.52 & 0.50 & 0.50&0.53\\
    \midrule
    \multirow{1}{*}{Single Attack}
    & - & 0.45 & 0.53 & 0.53 & 0.58 & 0.85 & 0.68\\
    \midrule
    \multirow{6}{*}{\shortstack{Co-membership Attack}}
    & 2 & 0.47 & 0.58 & 0.52 & 0.56 & 0.88 &  0.74 \\
    & 4 & 0.46 & 0.63 & 0.57 & 0.70 & 0.95 & 0.79\\
    & 8 & 0.56 & 0.81 & 0.61 & 0.69 & 0.96 & 0.76 \\
    & 16 & 0.60 & 0.70 & 0.71 & 0.79 & - & -\\
    & 32 & 0.68 & 0.82 & - & - & - & -\\
    & 64 & 0.92 & 0.94 & - & - & - & -\\
    \bottomrule
  \end{tabular}
  \vspace*{-2mm}
\end{table*}
\section{Experimental Results}
\label{sec:experiment}
\subsection{Evaluation of Membership Attack}
In this sub-section, we evaluate the effectiveness of the proposed (co)-membership attackers and show that some popular datasets with common network architecture and training methods are susceptible to such attackers. It exposes the unexpected privacy risk of publishing generative models even when they were trained with large datasets.

\smallskip\noindent\textbf{Comparison study.}  We tested the membership attacks on three datasets, MNIST~\cite{MNISTDataset}, CelebA and ChestX-ray8~\cite{wang2017chestx}. All three datasets are trained by using only the image feature $\x$. 
After the generative models are trained on the private training data, only the generative component is exposed to the attacker. To test the effectiveness of the attackers, we create an evaluation dataset by mixing training and non-training data (with their membership labels hidden from the attackers). Therefore, if an attacker is able to differentiate training and non-training instances among the evaluation dataset, the membership labels are recovered and privacy is leaked. For example, the experiment on MNIST uses 60,000 images as training and 10,000 as non-training data. For an attack instance, the adversary is randomly assigned only with a target image. It is stressed that the proposed attacker does not require any prior image from the training data or assumption about the underlying distribution. Then the attacker (reconstruction) loss on each instance is recorded. If the loss is smaller than a threshold, this target image is declared to be from the training dataset by the attacker. The ROC curve of such binary classifier is reported in Table~\ref{attack-table} by varying the discriminating threshold.


For references, we compare two other unsupervised methods with ours. One is a naive baseline method, named \emph{Nearest Neighbor}, in which we simply compare the given target instance $\x$ with the nearest neighbor in a set of generated data ($60000$ images for MNIST, $20000$ for CelebA and $1000$ for ChestX-ray8). The minimum L2-distance is considered as the attacker loss for this baseline attacker. The second compared baseline is called \emph{Direct Projection} (as in DefenseGAN~\cite{defenseGAN}), in which we take the noise $\z$ as trainable variable and use gradient descent to adjust the noise directly using the new objective $\min_{\z} \Delta(\x, G_\theta(\z))$. This direct method is not able to launch a co-attack since it is not possible to share information across different instances. 

To create the evaluation dataset of MNIST, $512$ images are randomly sampled from training and non-training data. 
Both baseline methods and the single attack treat each image as independent attack instances. While for 
co-attacks of strength $n$ 
the training (also non-training) images are partitioned into $512/n$ groups. Each group of $n$ images is considered as a single attack instance and co-membership attacks are launched against each instance.

\smallskip\noindent\textbf{Collaborative Machine Learning.}
For CelebA and ChestX-ray8, each image has an identity attribute which indicates the owner of that image. Each person may contribute multiple images to the dataset. This can be formulated in a collaborative learning setting: an aggregator initiates the learning process of a generative model by requesting users to contribute multiple images. Each user may choose to participate in the training or not. User's involvement (membership label) is private information which the attacker is trying to recover by accessing the trained model and the $n$ images (co-attack strength $=n$) of that user. For CelebA, 500 celebrities with at least 25 face images are randomly selected as contributors for the training data and for ChestX-ray8, 100 random patients with at least 10 scans contribute to the training data. Membership attacks are launched against 64 evaluation users sampled from the contributors and non-contributors. Therefore, the maximum co-membership strength is 20 and 10 for CelebA and ChestX-ray8 respectively. Table~\ref{attack-table} reflects this constraint.

\begin{figure*}[ht]
\begin{center}
\centerline{\includegraphics[width=1.8\columnwidth]{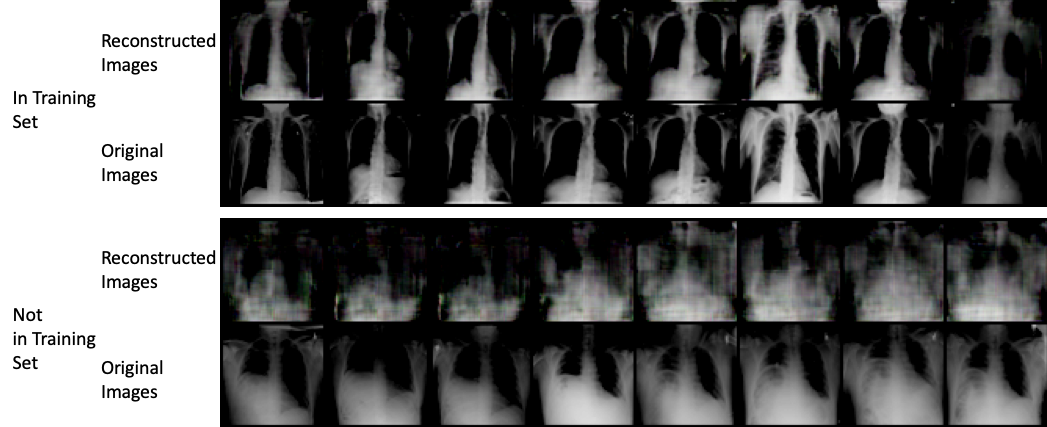}}
\caption{Visualization of generated results on ChestX-ray8}
\label{fig:chest}
\end{center}
\end{figure*}

    

\subsection{Implementation Details}

\begin{table}[ht]
  \caption{Network architecture used for MNIST}
  \label{net-table}
  \centering
  \begin{tabular}{lll}
    \toprule
    \multicolumn{2}{c}{WGANs / VAEs}  \\ 
    \cmidrule(r){1-2} \cmidrule(r){3-3}
    Generator/Decoder     					& Discriminator/Encoder \\ 
    \midrule
    FC $1024$ 							& Conv [4$\times$4], stride=$2$, 64  	\\ 
    FC $7\times7\times128$     				& Conv [4$\times$4], stride=$2$, 128	 \\ 
    Trans. Conv [4$\times$4], stride=$2$,$64$ 	& FC $1024$ 					 \\ 
    Trans. Conv [4$\times$4], stride=$2$,$1$	& 							\\
    \bottomrule
  \end{tabular}
\end{table}

The network architecture used in MNIST experiments can be found in Table~\ref{net-table}. All experiments in CelebA use the network architecture in DCGAN~\cite{radford2015unsupervised}. The ChestX-ray8 uses DCGAN for VAEs and DCGAN-like network with residual blocks for GANs. For the hidden layers in the neural networks, the activation function is ReLU and batch normalization is applied except the output layer of the generator/decoder. The last layer of the discriminator is a fully connected layer that maps the previous layer to a single value and then applies a sigmoid function. For the encoder, re-parameterization trick is used and the last layer outputs the mean and standard deviation of the Gaussian distribution for sampling. L2 regularization is applied to weights and biases of all hidden layers. Adam Optimizer is used with learning rate $0.001$ for the generator and $0.0001$ for the discriminator.

The model architecture of the attacker is relatively straightforward. The attacker network contains two fully connected hidden layers with 100 units and the final layer outputs the latent representation $\z$. Similar to the previous models, ReLu is chosen to be the non-linear activation function.  We run the attacker optimization (gradient descent) step for 1000 iterations and report the final L2-distance between the target and generated image as the reconstruction loss of the attacker. Since computing the (minimum) reconstruction loss in Eqn~\ref{eqn:attackerlossgan} is a non-convex problem, to reduce the effect of initialization we run the attack against the same instance for four times and take the minimum among the losses.

For the deep learning framework and hardware, TensorFlow is used to implement the deep neural networks. The training is done on an Nvidia Tesla K40c GPU.

\subsection{Attack Results} 
Our co-attacker outperforms  nearest neighbor and direct projection across all three datasets and for both VAEs and GANs.
The AUC of our method is able to reach around $75\%$ to $95\%$. The result shows the success of co-membership attack against deep generative models trained on popular datasets. To investigate the two factors behind the scene, namely training data size and co-membership information, we aim to understand when a membership attack succeeds. Then we report experimental results related to co-membership information. 

    
\begin{figure}[h]
\begin{center}
\centerline{
\includegraphics[width=0.825\columnwidth]{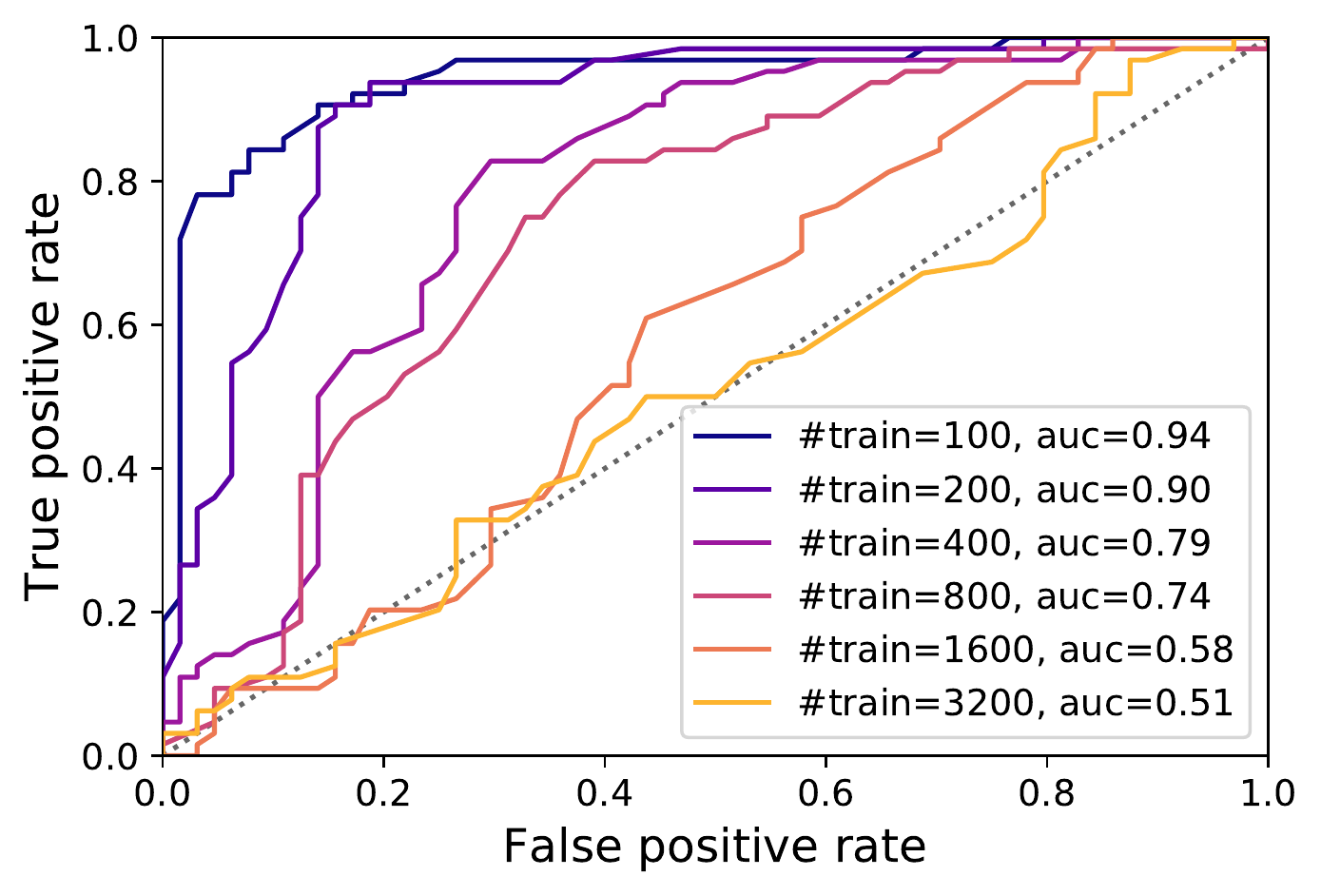}
}
\caption{ROC curve for single membership attacker against WGANs trained on MNIST with varying training data size.}
\label{fig:roc}
\end{center}
\vskip -0.2in
\end{figure}

\begin{figure}[h]
\begin{center}
\centerline{
\includegraphics[width=0.99\columnwidth]{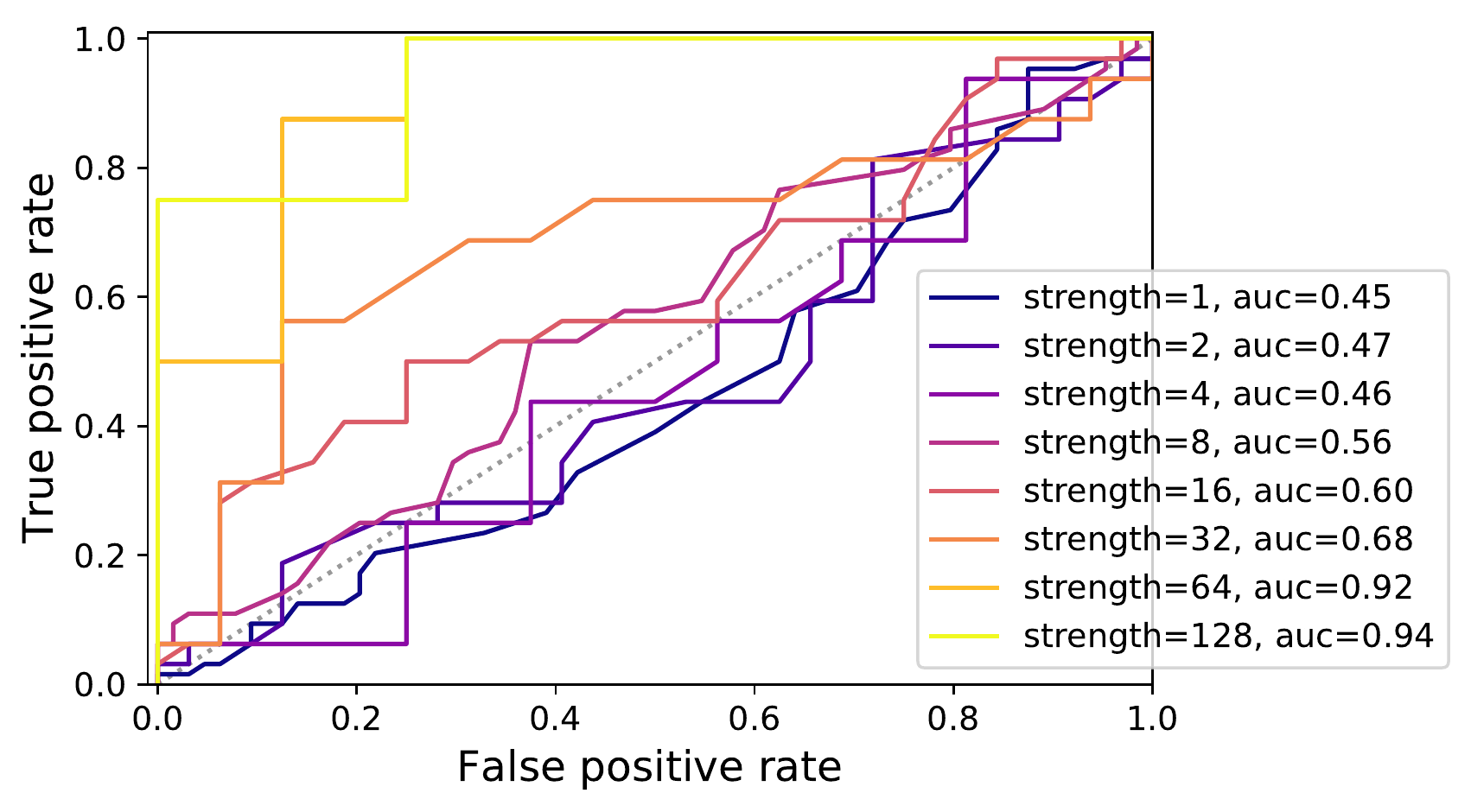}
}
\caption{ROC curve for co-membership attacker against WGANs trained on full MNIST with varying co-membership strength.}
\label{fig:roc_co}
\end{center}
\end{figure}

\smallskip\noindent\textbf{Single membership attack.} From Figure~\ref{fig:roc}, we observe that the ROC curve for the GANs trained with different training data size varies significantly. If the training dataset consists of only hundreds of images, the attacker can easily find a good latent representation for the generator to reproduce any target image from the training data. It suggests an overfitting scenario. 
This results in high attacker effectiveness for the simple binary classifier. However, when the training dataset size increases to thousands (orange and yellow curves), this difference vanishes gradually. Thus, the ROC curves are very close to the diagonal. Therefore, a single attacker may not have enough information to attack generative models trained with moderately sized dataset.


\smallskip\noindent\textbf{Co-membership attack.} While the generative models are safe against single attacks when trained on a large data set, 
Figure~\ref{fig:roc_co} shows that a privacy leakage exists if the attacker has the auxiliary co-membership information and utilizes the proposed co-membership attacker to fuse information together. The AUC of a co-membership attacker can even exceed 90\% for a model trained using the full-sized (60,000 images) MNIST dataset.

\begin{figure}[ht]
\begin{center}
\centerline{\includegraphics[width=0.85\columnwidth]{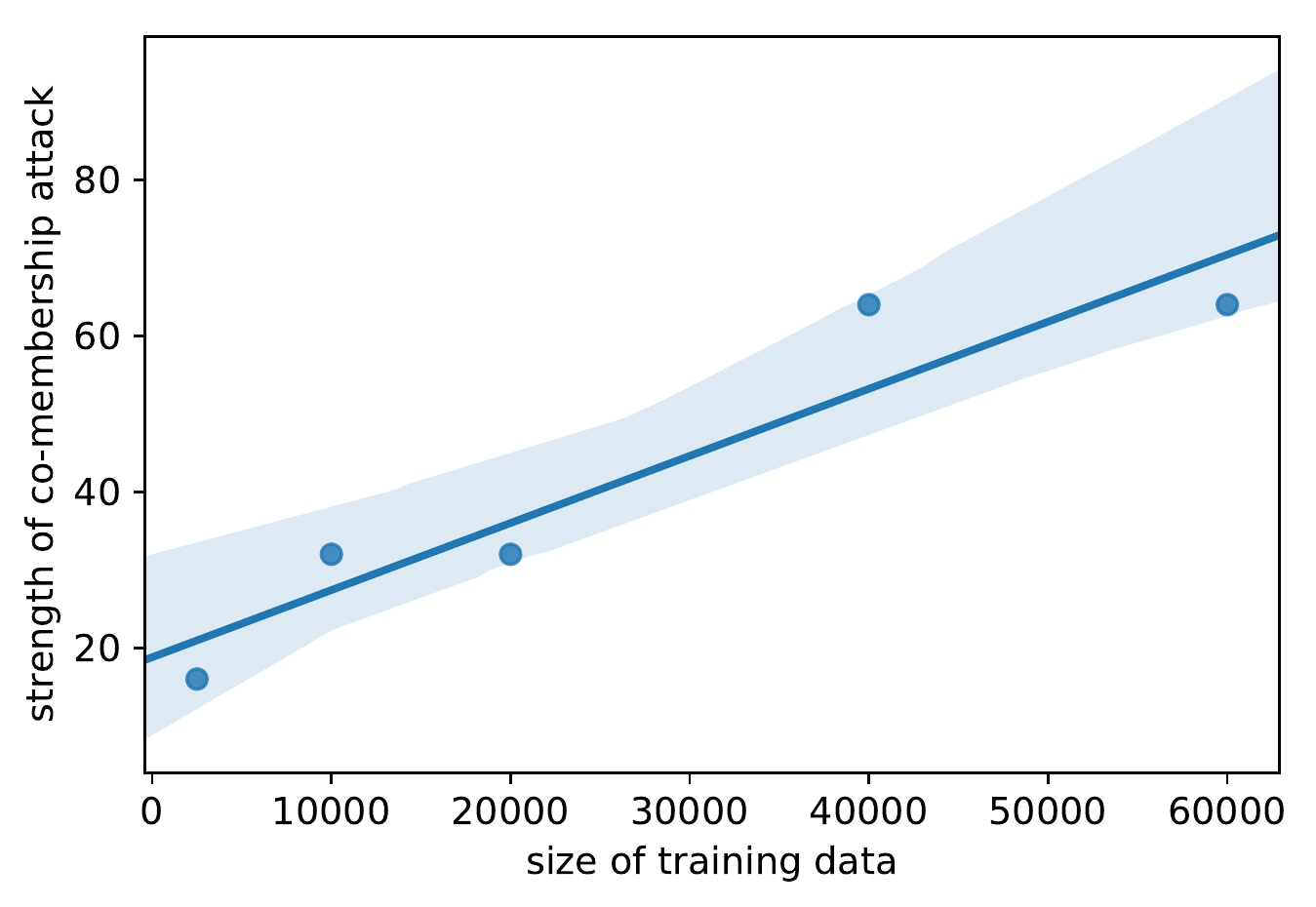}}
\caption{Relationship between minimum co-membership strength and training data size for an effective attacker (AUC$>$0.75)} 
\label{fig:reg}
\end{center}
\end{figure}

Figure~\ref{fig:reg} shows the relationship between the size of training data and the effectiveness of co-membership attacker on MNIST. 
Empirically, the figure suggests that $1000n$ training data points are needed for the model to be robust against co-attacks of strength $n$.

We also qualitatively visualize the generated results on ChestX-ray8 data in Figure~\ref{fig:chest}.
The first two rows show the generated and original images when the original (target) ones are actually contained in the training data. It is clear that the generated instances visually look similar with the original ones which indicates high attack/inference success rate. On the contrary, the lower two rows show similar instances while the original images are not in the training data. The results show that this time the generated instances are more random and far from the original ones.

\subsection{Relationship between Membership Attack and Model Generalization}
\label{sec:privacy_generalization}
Membership attacks for classifiers (supervised learning) were shown to be closely related to the generalization capability of the model~\cite{Yeom2017-qi}. 
As for generative models, there has not been an explicit discussion of how to relate model generalization with membership attacks. By providing a measure of generalization that is easy to compute, one can monitor the training process of a generative model and at the same time understand the privacy risk concerned with publishing the trained model.

Intuitively, we want a generalizing model to have similar training and non-training (testing) reconstruction loss -- the attacker is equally able to reproduce a target image from training or non-training data. Note that this generalization condition is only a necessary condition for distribution learning. A generator that is able to produce every element in the data distribution $\mu_{\real}$ (i.e., the generator `covers' all samples with non-zero measures in $\mu_{\real}$) cannot guarantee that it is generating the samples according to the distribution $\mu_{\real}$. 

Now, we make this intuition precise by measuring the \emph{generalization gap} of a generative model $G_\theta$ by the difference between the expected attacker loss on the training data (the finite samples) $\X$ and non-training data (testing data not used by the training procedure) $\mu_{\real} - \X$ which comes from the same underlying data distribution $\mu_{\real}$: 
\begin{equation}
\begin{split}
\underbrace{\mathbb{E}_{\x\sim\mu_{\real} - \X} \min_{\gamma} \Delta(\x,G_\theta(A_\gamma(\x)))}_{\text{attacker loss on non-training images}} - \\
\underbrace{\mathbb{E}_{\x\sim\X} \min_{\gamma}\Delta(\x,G_\theta(A_\gamma(\x)))}_{\text{attacker loss on training images}}
\end{split}
\end{equation}

\begin{figure}[ht]
    \centering    
    \begin{subfigure}[t]{\columnwidth}
    	\centerline{\includegraphics[width=0.85\columnwidth]{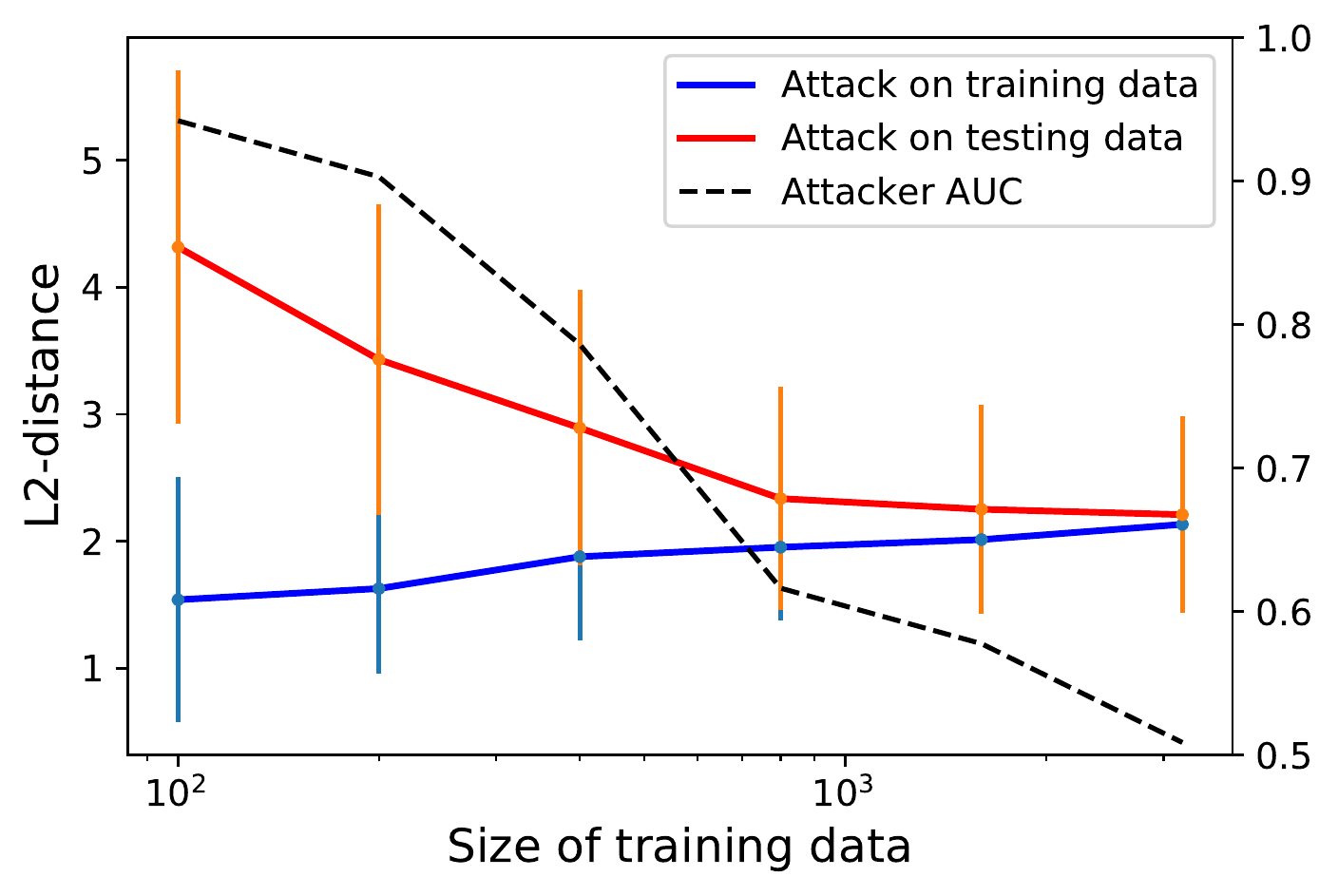}}
        \caption{Generalization gap and membership attack result. 
        The two attacker losses (with vertical bars showing one standard deviation) approach each other with more training data. Meanwhile, the success (AUC) of the membership attack is diminishing. }
        \label{fig:generalization}
    \end{subfigure}
    \vskip 0.2in
    ~
    \begin{subfigure}[t]{\columnwidth}
    	\centerline{\includegraphics[width=0.85\columnwidth]{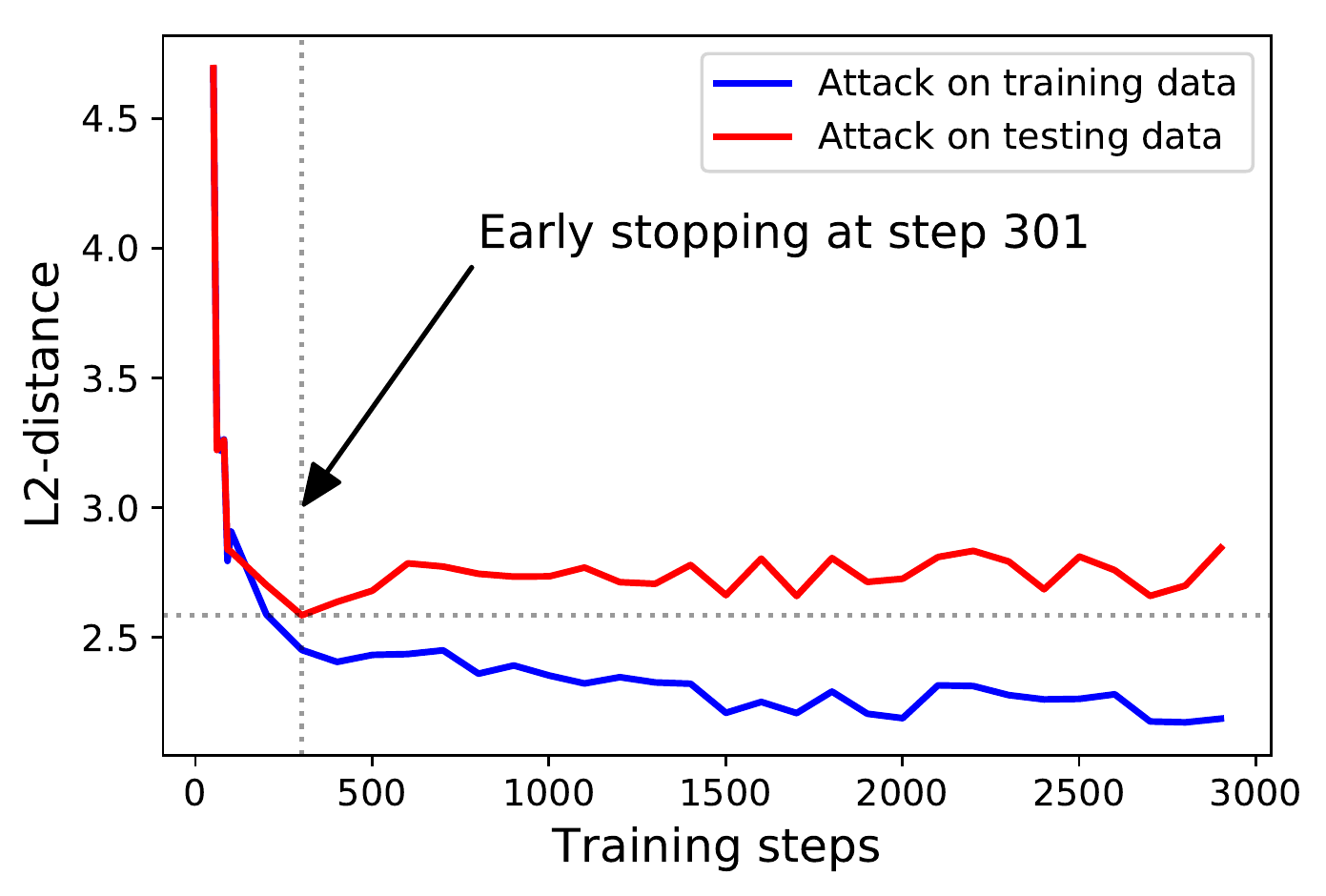}}
        \caption{Learning curve for WGANs trained with $800$ images. Model is over-trained after hundreds of training steps.}
        \label{fig:learning_curve}
    \end{subfigure}
    
    \caption{Generalization of WGANs on MNIST. The $y$-axis shows attacker loss in Eqn~\ref{eqn:attackerlossgan}. The blue curve and red curve show the average attacker loss for training data and non-training data respectively.}\label{fig:gan-generalization}
\end{figure}

\smallskip\noindent\textbf{Generalization Gap.}
In supervised learning, we should observe that for a given hypothesis class of classifiers, the generalization gap decreases when the number of training data increases. With the proposed measure of generalization for generative model, we find a similar pattern in Figure~\ref{fig:generalization}. For WGANs model trained on MNIST data and stopped after $2000$ training steps, the generalization gap diminishes when we use thousands of training data. We also plot the success rate of the single membership attacks. The success rate of membership attacks and generalization gap are strongly correlated.

\smallskip\noindent\textbf{Early Stopping.} On the other hand, the learning curve in Figure~\ref{fig:learning_curve} depicts how the training and testing errors behave when the number of training steps increases. For deep learning, the curve provides an early stopping scheme to prevent overfitting. As for the case of generative model, the training error also decreases steadily while the testing error increases again after some steps. This shows that generative model is easily overfitting (over-trained) when the dataset is small.


\newpage
\subsection{Diversity}

\begin{figure}[ht]
\begin{center}
\centerline{\includegraphics[width=0.85\columnwidth]{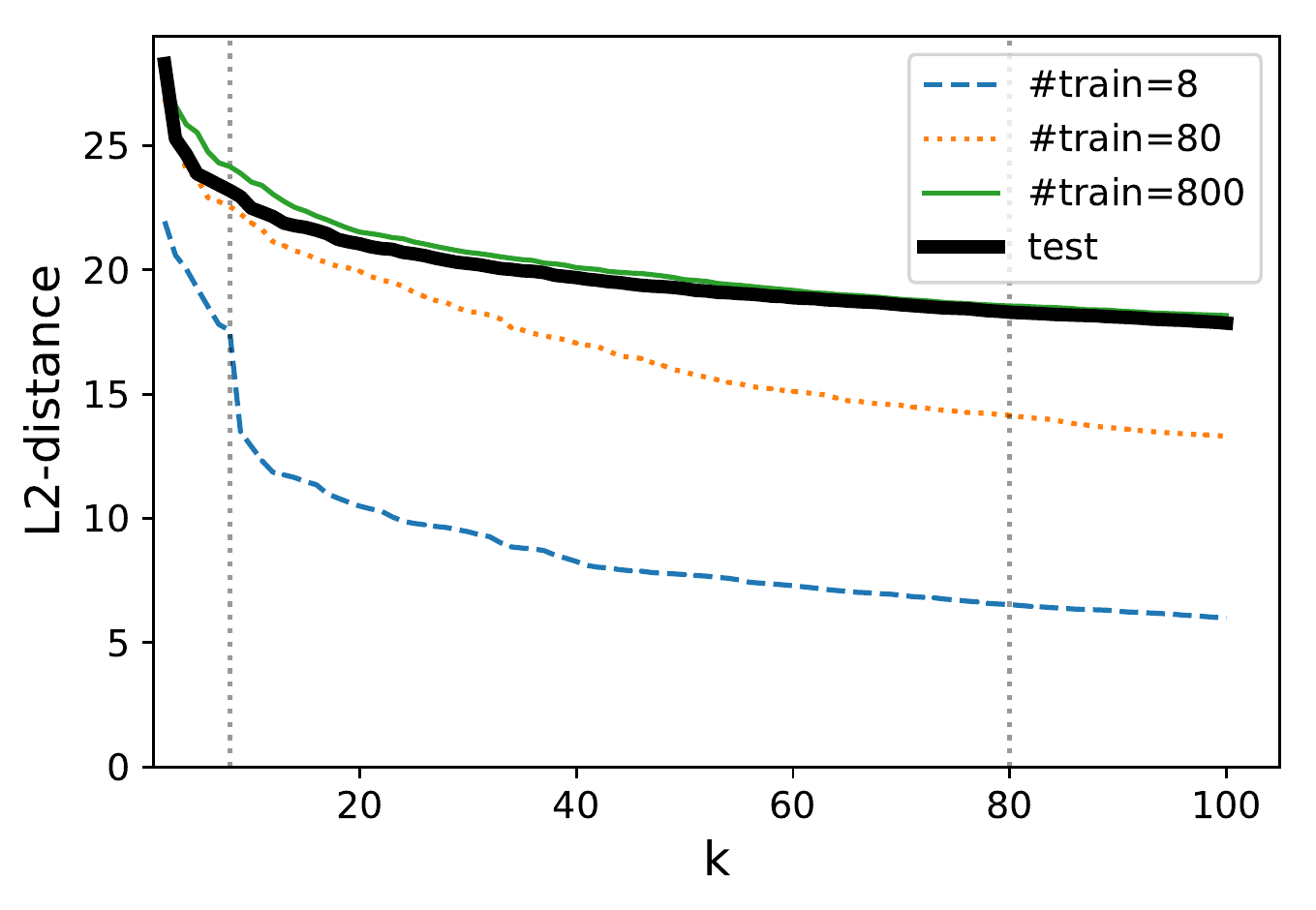}}
\vspace*{-2mm}\caption{$k$-dispersion of 3000 generated images from WGANs trained with different data sizes $=\{8,80,800\}$. `test' refers to the unknown ground truth data distribution which is represented by 10,000 images from the non-training portion of the image dataset as a proxy.}
\label{fig:dispersion}
\end{center}
\end{figure}

\smallskip\noindent\textbf{Dispersion.}
A desirable property of a generative model is the capability of generating versatile samples. This is termed \emph{diversity} in the literature. The evaluation of diversity, however, has not reached a consensus in the community. In particular, an obvious failing mode to avoid is that the GANs memorize the training data and simply report the training samples. There has been a number of proposed methods to test whether this is the case. For example, one test checks each generated image whether it is similar to any in the training data. Another test considers taking two random seeds $s_1, s_2$ and checks if the interpolation $\lambda s_1+(1-\lambda)s_2$ for $\lambda\in [0, 1]$ generates realistic outputs. Actually, no measure of diversity is explicitly defined. In a recent work~\cite{arora18GAN}, the authors have proposed to use a birthday paradox test: if the GANs have simply memorized the $n$ training sample and randomly output one each time, then with roughly $\sqrt{n}$ samples in the output there is a good chance that two of them are the same. It is then suggested in~\cite{arora18GAN} to visually examine the images to identify duplicates.  




\def\disp{\operatorname{disp}}

Here we use a geometric measure of diversity, called \emph{dispersion}, which seeks for a subset of $k$ images in the generated output set $\X_G$ that are far away from each other feature-wisely. 
$$\disp(k; \X_G) = \max_{\x_1,\ldots,\x_k \in \X_G} \min_{i,j \in [k]} \Delta(\x_i,\x_j)$$
If the output images are concentrated around a small number of $\ell$ samples, then the dispersion will dramatically drop when $k$ goes beyond $\ell$. In Figure~\ref{fig:dispersion}, we can observe such patterns in training GANs for MNIST dataset. For GANs trained with $8$ and $80$ images, they are not able to generate diversified images so that their dispersion is significantly lower than that of the testing (non-training) dataset which consists of 10,000 images in MNIST.

\smallskip\noindent\textbf{Connection with Generalization.} When a model generalizes, 
the dispersion of the generated data is similar to the dispersion of the ground truth data distribution. In Section~\ref{sec:privacy_generalization}, the generalization gap becomes smaller when we have more training data. So one can produce diversified data that are not seen in the training data. Data dispersion serves as an evidence for such diversification. When a generative model is successful, it should interpolate or even extrapolate beyond the original training data which results in a larger data dispersion.

\begin{figure*}[htbp]
\centering
    \begin{subfigure}[t]{0.25\textwidth}
        \includegraphics[width=\columnwidth]{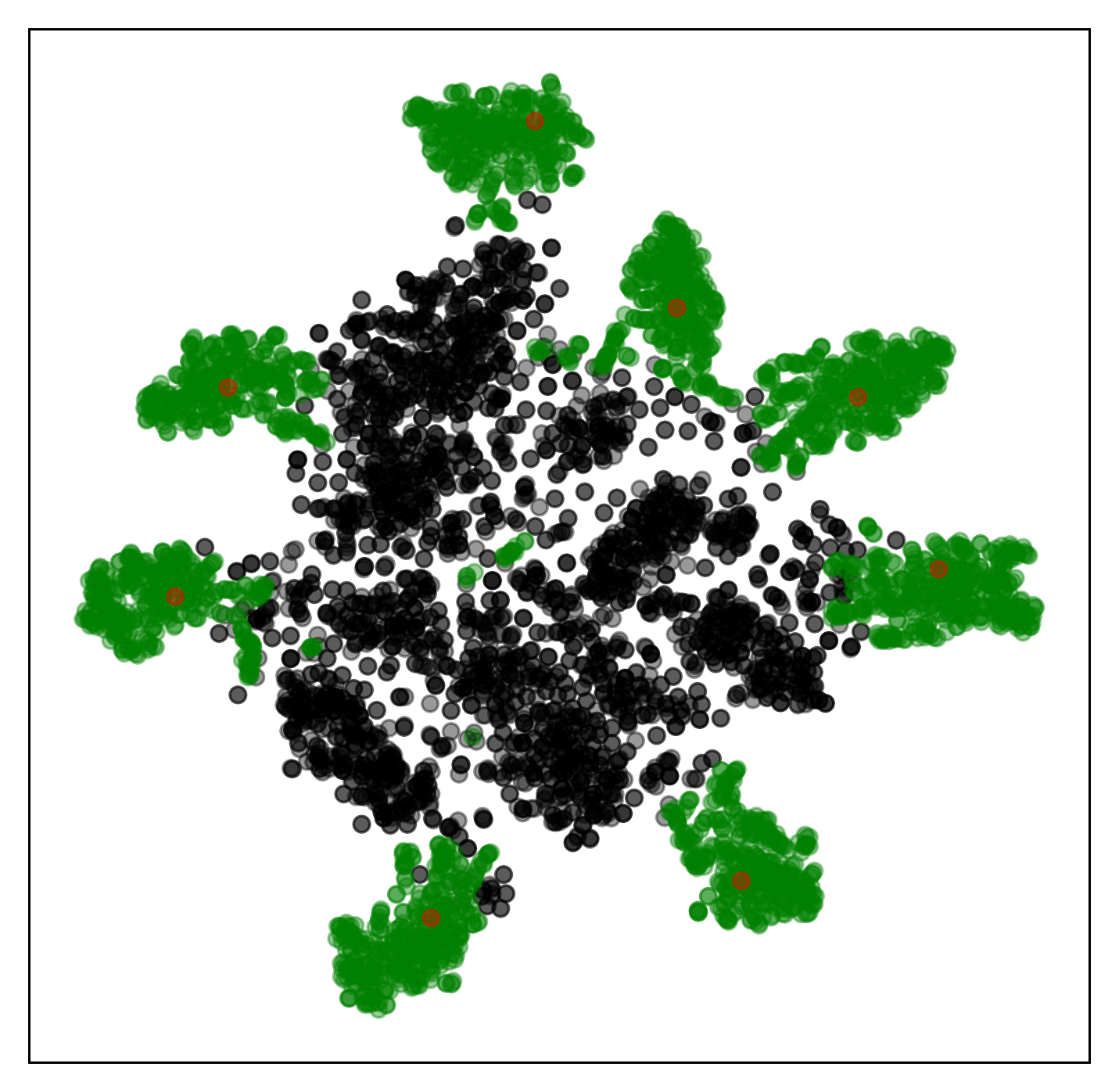}
        \caption{Training data size $=8$}
        \label{fig:tsne8}
    \end{subfigure}
    ~ 
    \begin{subfigure}[t]{0.25\textwidth}
        \includegraphics[width=\columnwidth]{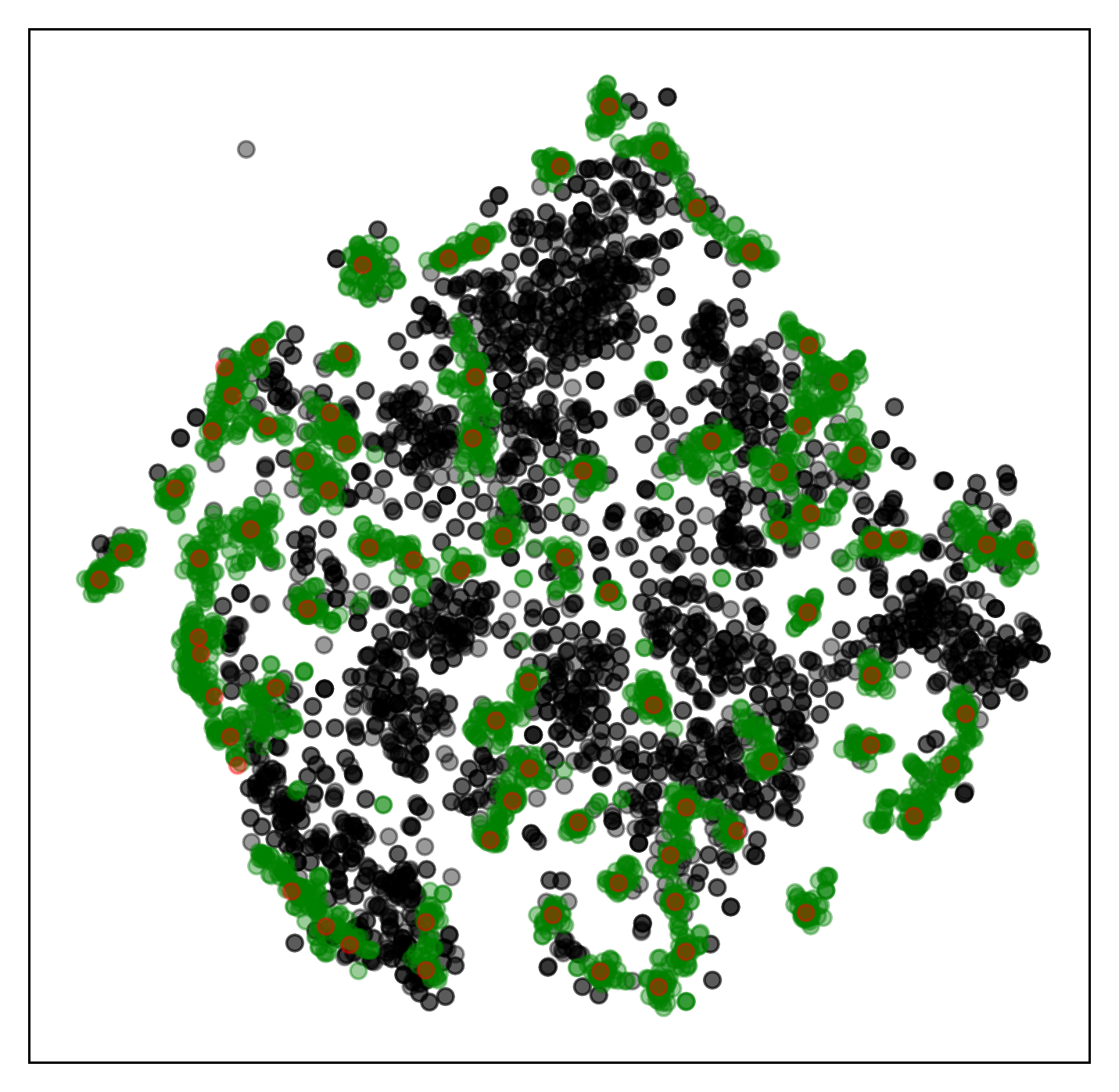}
        \caption{Training data size $=80$}
        \label{fig:tsne80}
    \end{subfigure}
    ~ 
    \begin{subfigure}[t]{0.25\textwidth}
        \includegraphics[width=\columnwidth]{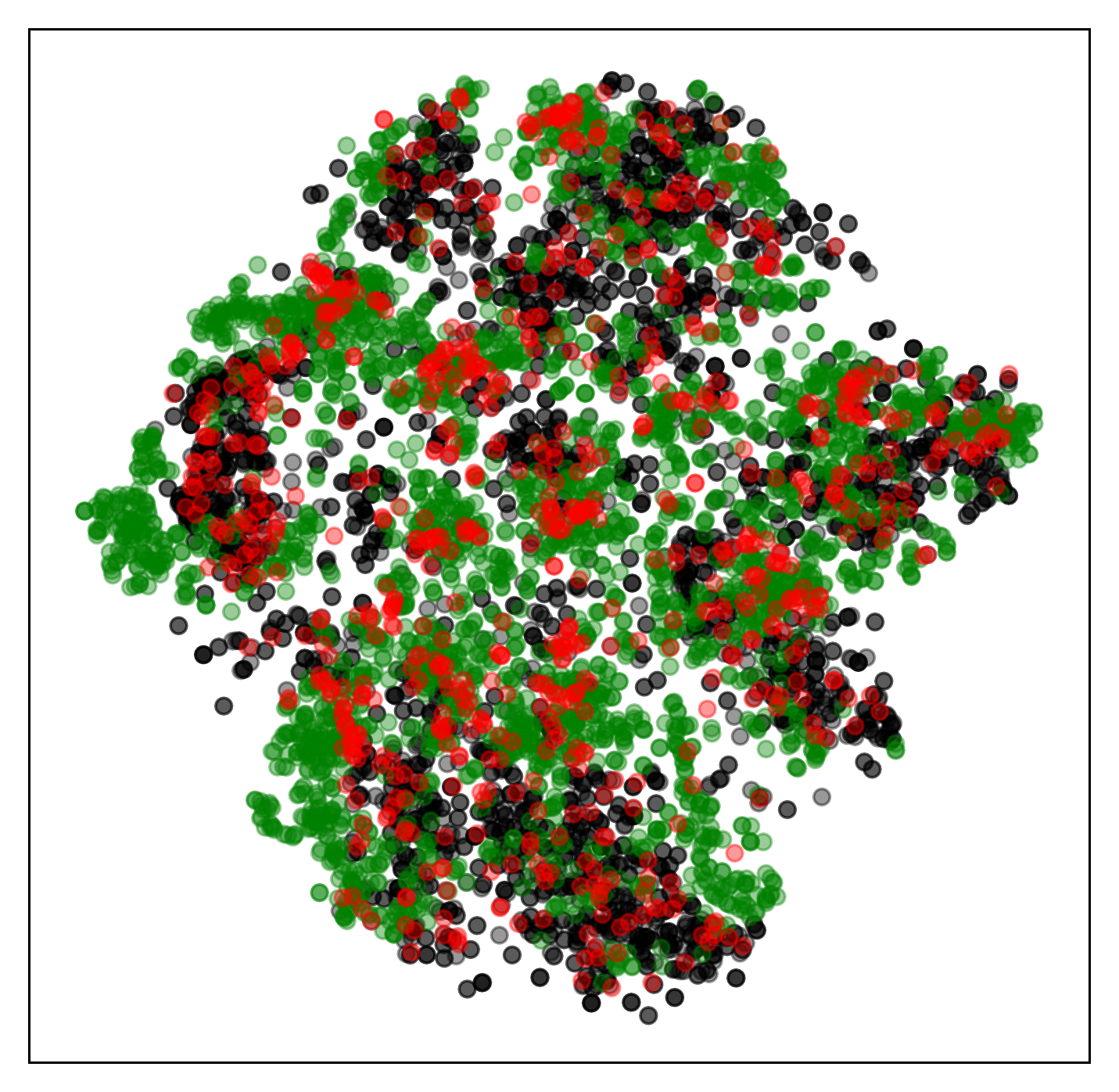}
        \caption{Training data size $=800$}
        \label{fig:tsne800}
    \end{subfigure}
    \caption{$t$-SNE visualization of training images (in red), generated images (in green) and images from real distribution (in black) for WGANs trained with different sizes of training data.}\label{fig:tsne}
\end{figure*}

In Figure~\ref{fig:tsne}, we visualize dispersion by embedding the generated images and the training data on a two-dimensional map using t-SNE \cite{Maaten2008-zd}. 
The red dots are the training data, the number of which is the only changing variable in the subfigures. The green dots show the $3000$ generated images sampled from the GANs after training. The black dots represent $3000$ samples from the real distribution. When there are only a small number of training images, the generated samples heavily concentrate around the training data which produces a small dispersion. Once the number of training data is $800$, the dispersion of the generated data and original data becomes similar. 



\subsection{Diversity versus Model Generalization}

\begin{figure}[htbp]
\centering
    \begin{subfigure}[t]{0.8\columnwidth}
        \includegraphics[width=\columnwidth]{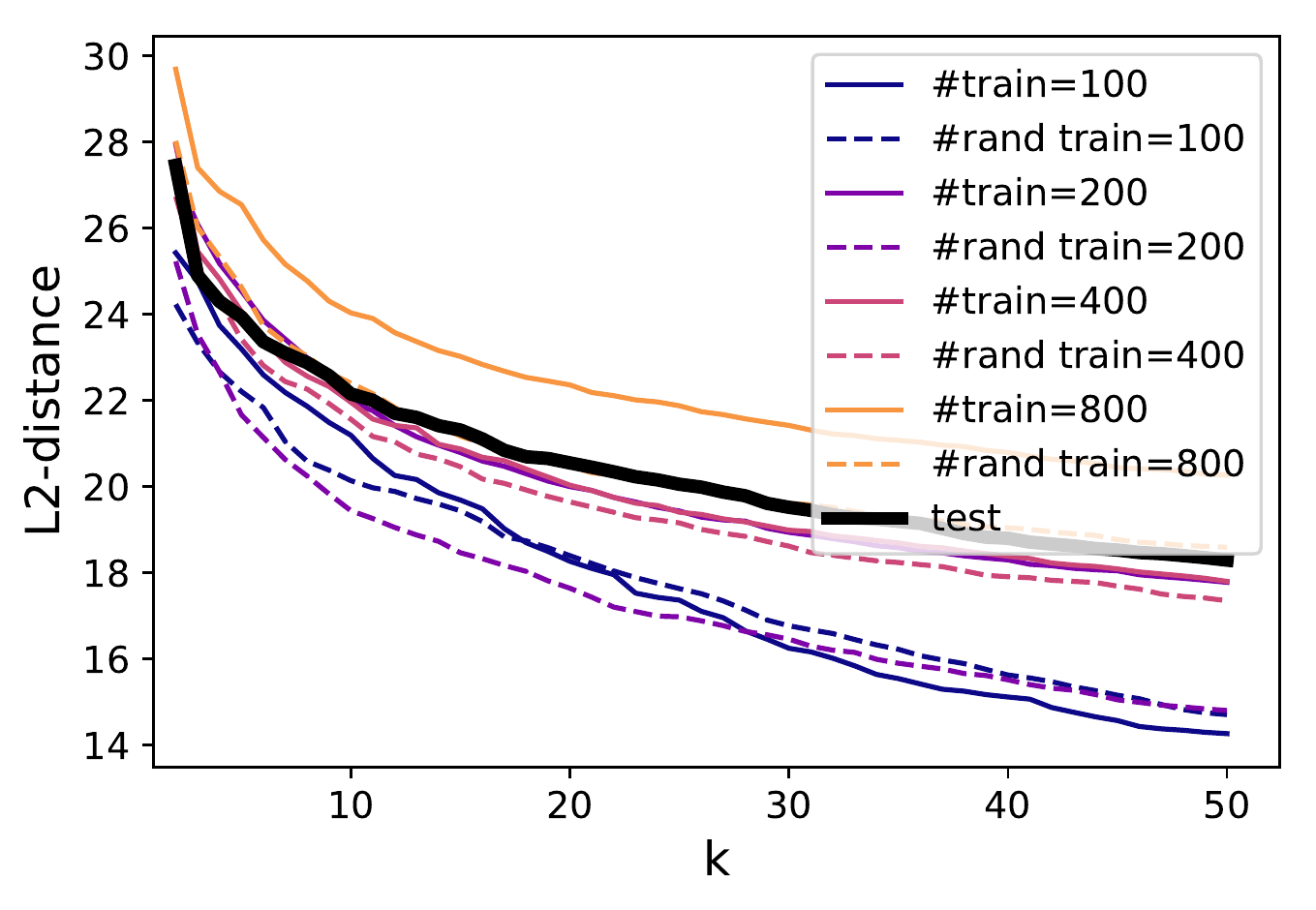}
        \caption{Dispersion with Adversarial Sampling} 
        \label{fig:dispersion_active}
    \end{subfigure}
    ~ 
    \begin{subfigure}[t]{0.8\columnwidth}
        \includegraphics[width=\columnwidth]{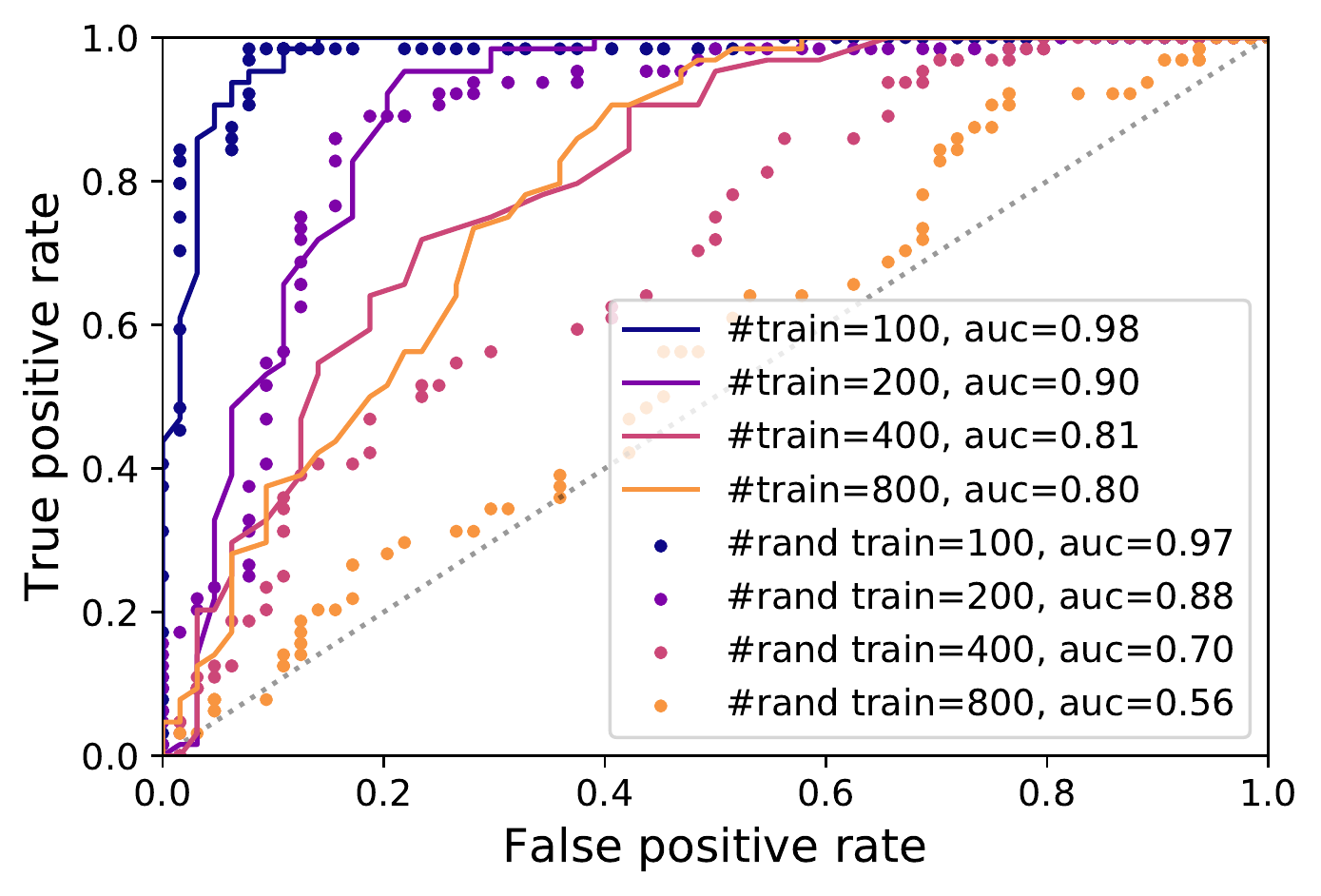}
        \caption{Membership Attack with Adversarial Sampling}
        \label{fig:roc_active_subsettest}
    \end{subfigure}
    \caption{Adversarial sampling versus random sampling for diversity and generalization. For adversarial sampling, dispersion is enhanced while membership privacy risk is increased. }\label{fig:diversity}
\end{figure}


It is interesting to ask whether diversity and generalization capabilities are the same. And it is tempting to conclude that a model with good diversity has good generalization. In this sub-section, we point out that these two measures (goals) are not always aligned. One could carefully choose training data so that diversity is enhanced yet generalization is hurt -- i.e., membership attacks are more likely to be successful. 


For example, when we take a batch of data that have not been used in the training process, $\{\x_i\}_{i=1}^b$, and 
assign a ranking to them by the decreasing order of the attacker loss (Eqn~\ref{eqn:attackerlossgan}) with respect to the current generator. We include the data point ranked highest (the one that is the least reproducible) in $\X'$. Then we modify the current GANs by training on all data in the current batch and repeat the above procedure. Eventually, we have a subset $\X'$ that is the hardest one to reproduce in the process.




After the above procedure has completed, a subset of data points $\X' \subset \X$ is collected from the original training dataset $\X$. Now we train a new GANs from scratch using data in $\X'$. In comparison, we also train a separate GANs using a set $\Y$ with $|\Y|=|\X'|$, randomly selected from $\X$ as the control group. We denote $\X'$ to be obtained by \emph{adversarial sampling} and $\Y$ as random sampling. 



Hereafter, we compare the dispersion and rate of success of membership attacks on GANs trained with adversarial sampling and uniform random sampling. The dispersion of generated data using the subset $\X'$ is even higher than the dispersion of the original training data (Figure~\ref{fig:dispersion_active}). But in terms of membership attack, GANs trained by adversarial samples are much worse (Figure~\ref{fig:roc_active_subsettest}). One way to understand this result is that the adversarial samples might have paid too much attention on extreme cases while random sampling is a better unbiased representation of $\mu_{\real}$.

\section{Conclusion}
To explore the privacy vulnerabilities of generative models, we propose co-membership attacks and compared with single membership attacks on several datasets. 
We show that different generative models are possible to leak sensitive membership information under various attack scenarios. We hope this work will encourage further privacy protection mechanisms for generative models.

\section*{Acknowledgment}
This work is partially supported through NSF DMS-1737812, NSF CNS-1618391, NSF CCF-1535900, DARPA FA8650-18-2-7882, and Wechat Graduate student fellowship.

\bibliographystyle{IEEEtran}
\bibliography{reference_gan,private-sensing,privacy,learning}

\end{document}